# Two novel feature selection algorithms based on crowding distance

Abdesslem layeb

LISIA lab., Computer science and its application department, NTIC faculty, university of Constantine 2
abdesslem.layeb@univ-constantine2.dz

**Abstract.**

In this paper, two novel algorithms for feature selection problem are proposed. The first one is a filter method while the second is wrapper method. Both the proposed algorithms use the crowding distance used in the multiobjective optimization as a metric in order to sort the features. The less crowded features have great effects on the target attribute (class). The experimental results have shown the effectiveness and the robustness of the proposed algorithms.

## 1. Introduction

Feature selection problem is well known problem in data mining field. It is used in the classification methods to reduce the number of the features in datasets. Formally, feature selection procedure selects a subset of P significant features from a whole set of N input features, with P<N preserving a good or better accuracy compared to the entire N features [1]. It should be noted that feature selection is different from dimensionality reduction. Dimensionality reduction methods create new features by combinations of attributes, whereas feature selection methods use a subset of the present attributes without changing them.

Several methods were developed to solve feature selection problem that can be regrouped in three general classes of methods: filter methods, wrapper methods and embedded methods.

In the filter feature selection methods, each potential feature is weighted and ranked according to a defined feature selection measure, the selection procedure consists to take the best k features. The filter methods select the potential feature independently to the classifier used. Some examples of some filter methods include the Pearson Correlation Coefficient [2], and relief feature selection [3].

Unlike filter feature selection methods, Wrapper methods consider the feature selection problem as an optimization problem, where an objective function based on a predictive model is used to assess the accuracy of each selected subset of features. In this class, we can find complete search methods like branch and bound [4] or incomplete search methods like local search methods, greedy search, or metaheuristics [5]. Generally, the wrapper methods give better results than the filter methods. However, there are two main drawbacks that limit the use of these methods. the first limit is the runtime complexity required for the selection. The second limit is that the performance of these methods depends on the classifier algorithm used as objective function.

Finally, the embedded methods integrate the feature selection in the construction process of prediction model. In wrapper type selection methods, the classification process is divided into two parts: a learning stage and a validation stage to validate the selected subset of features. On the other hand, the built-in methods can use all the learning examples to build the system. This is an advantage that can improve the results. Another advantage of these methods is their speed compared to wrapper approaches because they avoid the classifier to be restarted for each subset of characteristics. Examples of embedded algorithms are the LASSO, Elastic Net and Ridge Regression [6].

As we have mentioned, the filter methods depend greatly on the metric used to assess each feature. In this work, we proposed the use of a new metric for ordering the features based on the famous crowding distance used in the multiobjective optimization [7]. The features are handled as points in multiobjective space where the objectives are the samples. The crowding distances of the entire features are sorted in descending order. Consequently, the selected features are ranked in the top of the features ranking.

The proposed algorithms are assessed with well-known datasets and they were compared against well-known algorithms. The experimental results have proved the effectiveness of the proposed algorithms.

In the following, we will explain in more details the proposed algorithm and the experimental study.

## 2. The proposed feature selection methods based crowding distance

In this work, two feature selection algorithms based on crowding distance are proposed. the first algorithm is a filter method while the second is wrapper method. Both the two algorithms use the crowding distance to sort the features. The use of the descending crowding distance is motivated by the following assumption that the most isolated features have great impacts on the target feature (class) while two closed (or crowded) features A and B have a close impact on the target feature. Therefore, it is preferably to select first the most isolated features than the most crowded features. The crowding distance is adapted as follows:

First of all, the features are sorted according to all the sample $S_m$ where a sample $S_m$ plays the role of an objective function in multiobjective optimization problems. The vectors of sorted indices $\mathbf{I}_m$ are found. The crowding distance $CD$ for each feature is computed using the following equations:

$$CD(I_m(i)) = \sum_{m=1}^{M} CD_m(I_m(i)) \quad (1)$$

Where

$$CD(I_m(i)) = \frac{Sm(I_m(i+1)) - Sm(I_m(i-1))}{S_{m,max} - S_{m,min}} \quad (2)$$

where $I_m(i)$ is the *i*-th index from the *m*-th vector of indices, $S_{m,max}$ and $S_{m,min}$ are the maximal and minimal values of the *m*-th Sample data, respectively. The value $CD_m$ for the two extreme features is set to infinity. Geometrically, the crowding distance is the average side length of the cuboid defined by features surrounding a particular feature (see Fig. 1). The less crowded features with a great value of $CD$ are the preferred features.

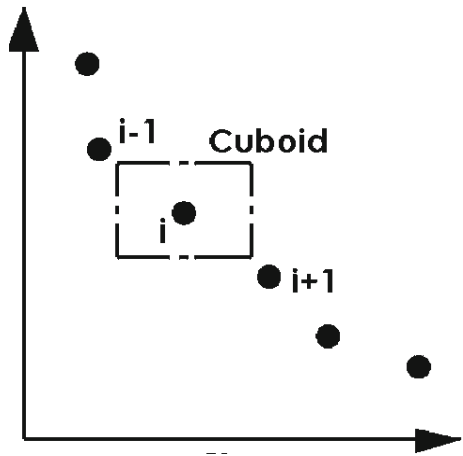

**Figure 1.** Crowding distance.

### 2.2 Filter algorithm based crowding distance

Figure 2 displays the outline of the proposed filter algorithm based crowding distance. After reading the dataset, the algorithm computes the crowding distance for every feature, then the obtained distances are sorted in descending order. Finally, the user selects the k first features in the ordered features vector.

```
    1. Read X=dataset with m samples and n features
    2. CD=Compute the crowding distance of n features
    3. Orderedfeatures =sort (CD, descending)
    4. Feats=Select the k first features in the ordered features
Compute the accuracy of the reduced dataset  X(Feat) by using a certain classifier
```

**Figure 2.** the pseudo code of the Filter crowded features

### 2.3 Wrapped algorithm based crowding distance

The wrapped algorithm is based on greedy method where at each step one feature is added to the selected features, if the accuracy of the classifier is enhanced then we keep this feature otherwise it is discarded. The choice of the feature to be added is given by the order of the features computed by the crowding distance. At each step, the fitness of the current solution is computed by a given classifier. Moreover, we can add other termination criteria to stop the algorithm early like accuracy threshold. The outline of the proposed algorithm is given in the figure 3.

```
    1. Read X=dataset with m samples and n features
    2. CD=Compute the crowding distance of n features
    3. Orderedfeatures=sort (CD, descending)

    4. For i=1 to n

       Selectedfeatures= Selectedfeatures U { Orderedfeatures (i)}
       Fitness=Compute the accuracy of the selected features by classifier F
       If Fitness >Bestaccuracy
       Update the bestaccuracy
       Save the selectedfeatures;
       Else
       Delete the last added features:
       Selectedfeatures / { Orderedfeatures (i)}
     end
```

**Figure 3.** the pseudo code of the wrapped crowded features

## 3. Implementation and results

The proposed filter and wrapped feature selection algorithms based on crowding distance are implemented under MATLAB R2016a environment, and all experiments were carried out on a Windows 10 64-bit computer with an Intel i3 (2.3 GHz) processor and 4 GB RAM. To evaluate the proposed algorithms, six popular datasets were used. The details about the used datasets are described in the table 1. Due to the randomness of the k-fold cross-validation, for one dataset, each algorithm is executed 30 times, and the best, mean, std, and worst results are reported. For all the algorithms, the multiclass SVM classifier is used with kfold=5. For the filter algorithms, the number of selected features is fixed to 10 for Ionosphere, Breast, Heart, and Sonar datasets, while for large datasets like Ovarian and Colon datasets, the number of selected features is fixed to 150.

**Table 1 :** Experimental Datasets

| Dataset | Data Type | # features | # samples |
|---|---|---|---|
| **Ionosphere** | Radar Data | 34 | 351 |
| **Breast** | Breast cancer Data | 30 | 569 |
| **Heart** | Heart deseas | 44 | 267 |
| **Sonar** | Signals Rocks vs Mines | 60 | 208 |
| **Ovarian** | Ovarian cancer Data | 4000 | 216 |
| **Colon** | Colon cancer Data | 2000 | 62 |

Table 2 shows the experimental results found by our filter algorithm called Filter Crowded Features. Moreover, our algorithm is compared to the most popular filter algorithms: Pearson Correlation Coefficient [2], Relief Feature [3] and Variance Feature Selection [8].

**Table 2.** the experimental results of filter algorithms for feature selection

| dataset | # features selected | Filter Crowded Features | | | | Pearson Correlation Coefficient | | | | Relief Feature | | | | Variance Feature Selection | | | |
|---|---|---|---|---|---|---|---|---|---|---|---|---|---|---|---|---|---|
| | | mean | std | worst | best | mean | std | worst | best | mean | Std | worst | best | mean | std | worst | best |
| **Ionosphere** | 10 | 94.13 | 0.73 | 92.31 | 95.16 | 92.19 | 0.68 | 90.60 | 93.16 | 95.23 | 0.42 | 94.29 | **96.02** | 89.07 | 0.52 | 88.03 | 90.04 |
| **Breast** | 10 | 92.52 | 0.39 | 91.92 | 93.31 | 92.37 | 0.33 | 91.74 | 93.32 | 92.58 | 0.37 | 92.09 | 93.49 | 92.53 | 0.24 | 92.09 | 93.15 |
| **Heart** | 10 | 78.25 | 0.60 | 76.76 | 79.40 | 79.60 | 1.10 | 76.78 | 81.64 | 79.05 | 1.40 | 76.04 | 82.01 | 79.62 | 1.50 | 76.42 | **82.77** |
| **Sonar** | 10 | 71.53 | 1.97 | 67.31 | 74.58 | 75.83 | 1.29 | 73.54 | 78.34 | 83.45 | 1.52 | 80.73 | **86.05** | 77.62 | 1.68 | 73.53 | 81.73 |
| **Ovarian** | 150 | 94.52 | 0.75 | 92.12 | 95.43 | 89.64 | 1.15 | 86.56 | 91.66 | 94.62 | 0.85 | 93.04 | 96.30 | 95.26 | 0.69 | 93.99 | **96.74** |
| **Colon** | 150 | 82.38 | 1.90 | 77.56 | 85.64 | 83.02 | 2.29 | 78.97 | **87.31** | 81.39 | 1.57 | 77.56 | 83.97 | 81.40 | 1.43 | 78.97 | 84.10 |

From the table 2, we observe that there is no great difference between our algorithm and the other algorithms. Our algorithm gives weak accuracy in sonar dataset and it gives accuracy close to the other algorithms in the reminder datasets. Indeed, the Wilcoxon test shows that there is no significant difference between the proposed algorithm and the other algorithms at level 0.05.

The table 3 reports the experimental results of the wrapped version. As we can see, the proposed algorithm is able to find a small set of features with higher accuracy. In all the datasets, the best accuracy is greater than 84% and all the results are better than those of the filter algorithms (table 2). The importance of the crowding ordering is obviously clear in the case of ovarian and colon datasets where the number of the selected features is 27 over 4000 features for ovarian dataset and 23 over 2000 features for colon datasets.

**Tables 3.** Statistical results of the wrapped crowded features algorithm

| | # features for best accuracy | mean | std | worst | best |
|---|---|---|---|---|---|
| **Ionosphere** | 16/34 | 94.87 | 0.80 | 93.15 | **96.30** |
| **Breast** | 8/30 | 93.15 | 0.44 | 92.09 | **94.91** |
| **Heart** | 12/44 | 81.64 | 1.33 | 79.40 | **84.63** |
| **Sonar** | 16/60 | 86.32 | 1.48 | 83.17 | **89.38** |
| **Ovarian** | 27/4000 | 93.23 | 2.39 | 87.04 | **96.78** |
| **Colon** | 23/2000 | 85.45 | 2.70 | 79.36 | **91.92** |

4. **Conclusion**

In this paper, two algorithms for feature selection are presented. The main feature of the proposed algorithm is the use of the crowding distance to order the features from the most important to the less important. The first algorithm is a filter method whereas the second is wrapped method. The experimental results show the effectiveness of the proposed algorithms compared to most popular feature selections algorithms. Besides, the wrapped version can be improved by introducing more specific operators.